\documentclass[runningheads]{llncs}

\usepackage{eccv}

\usepackage{eccvabbrv}

\usepackage{graphicx}
\usepackage{booktabs}

\usepackage[accsupp]{axessibility}  

\usepackage{hyperref}

\usepackage{orcidlink}
\usepackage{algorithm}
\usepackage{algpseudocode}
\usepackage{tabularx}
\usepackage{multirow}
\usepackage{makecell}
\usepackage{pifont}
\usepackage{float}
\usepackage{anyfontsize}
\usepackage{marvosym}

\begin{document}

\title{HitMem: Hierarchical Temporal 3D Memory with Multi-Modal Context-Aware Retrieval \\ for Dynamic Environments} 

\titlerunning{HitMem}

\author{Ruijie Tang\inst{1,2,4} \and
Chenye Zou\inst{3} \and
Guoquan Wu\inst{1,2,4}\textsuperscript{(\Letter)} \and \\
Jun Wei\inst{1,2,4} \and
Wei Chen\inst{1,2,4} \and
Jiaxin Zhu\inst{1,2,4}}

\authorrunning{R. Tang et al.}

\institute{Institute of Software, Chinese Academy of Sciences (ISCAS), Beijing, China \and
University of Chinese Academy of Sciences, Beijing, China \and
Alibaba Group, Shanghai, China \and 
Beijing Key Laboratory of Intelligent Software Engineering, China }

\maketitle

\begin{abstract}
Executing long-term tasks in dynamic environments requires embodied agents to maintain robust and adaptive 3D scene representations. However, most existing 3D memory frameworks rely on static world assumptions. When objects are displaced by human activities or unobserved events, agents encounter memory-observation conflicts and often require costly geometric recomputations or inefficient global re-exploration.
To address this, we propose HitMem, a hierarchical temporal 3D memory framework with a multi-modal context-aware retrieval mechanism. 
Through continuous perception, HitMem unifies semantic and spatial information into a lightweight topological graph that captures support relationships, while a temporal decay mechanism dynamically regulates memory activeness to mitigate the impact of stale representations. 
In addition, the multi-modal context-aware retrieval mechanism defaults to filtering candidates using integrated semantic, spatial, and temporal memory features, and activates a specialized two-stage retrieval process when object displacement is detected. This process combines spatial constraints inferred from external agent trajectories with semantic common sense grounded in class affinities, efficiently identifying high-probability candidate regions.
Extensive evaluations on our constructed Dyna-THOR benchmark demonstrate that HitMem significantly improves object relocation accuracy, reduces exploration costs, and enhances task execution performance in dynamic environments.


\end{abstract}    
\section{Introduction}
\label{sec:intro}

\begin{figure} [tb]
    \centering
    \includegraphics[width=0.85\linewidth]{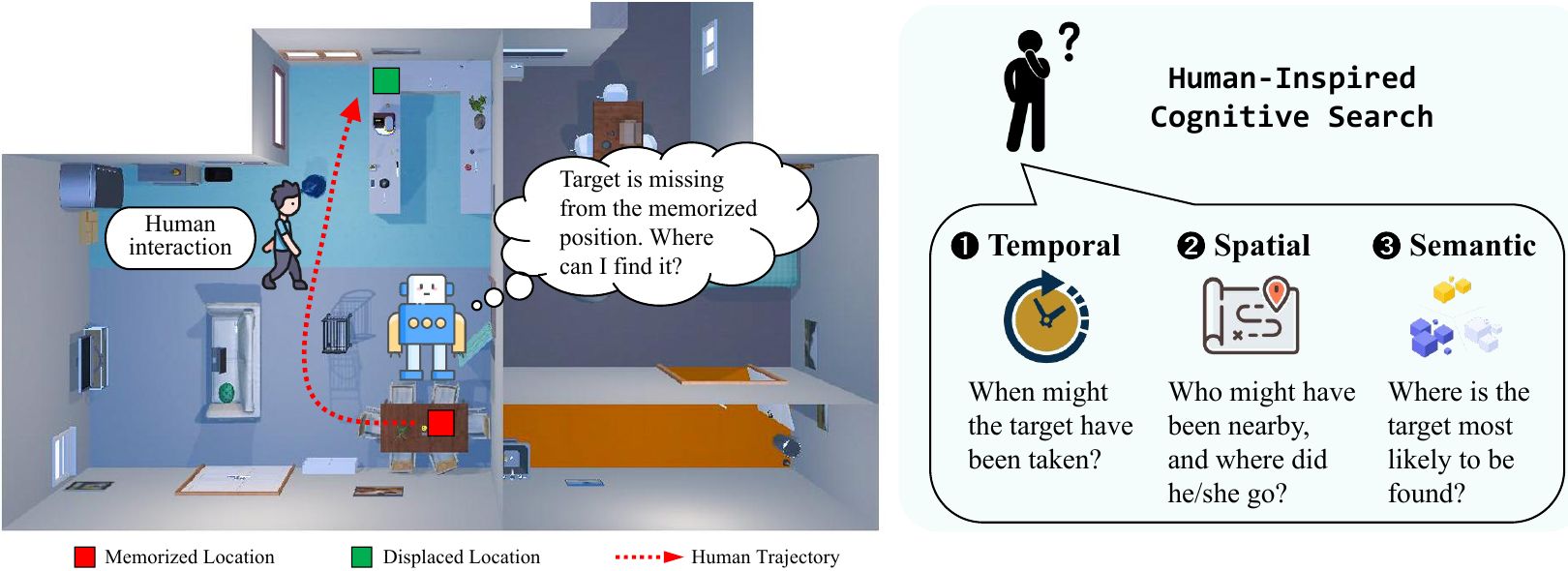}
    \caption{\textbf{Design intention of HitMem.} Existing methods are constrained by static environment assumptions or resort to blind exploration when targets are displaced. In contrast, we construct a hierarchical temporal memory and leverage multi-modal environmental information for retrieval, effectively enhancing adaptability to dynamic environments and improving exploration efficiency.}
    \label{fig:intention}
\end{figure}

Performing complex and long-term tasks within real-world environments is a fundamental objective of modern embodied agents. Such agents must operate continuously, respond to user instructions, autonomously identify targets, and execute corresponding actions \cite{yenamandra2024homerobotopenvocabularymobilemanipulation, zhang2024gamma, wu2025momanipvla}. A cornerstone of achieving these objectives is the integration of memory, which enables the agent to store and effectively retrieve environmental information to support task execution. Many recent methods have successfully transformed sensory inputs into 3D memories via dense reconstruction or scene graph modeling, equipping embodied agents with the requisite spatial awareness for navigation and manipulation \cite{huang2023voxposer, kerr2023lerf, mazur2023feature, lingelbach2023tasksg, yin2024sg-nav, maggio2024clio}.

However, existing methods are fundamentally limited by their reliance on a static world assumption. Real-world environments are intrinsically dynamic, where objects are frequently displaced by human activities or unobserved events. This discrepancy causes memory conflicts when an agent navigates to a location and finds the target missing. Current memory frameworks offer limited support for dynamic updates. Specifically, the re-inference of relational structures or the recomputation of geometric clusters is both computationally inefficient and prone to errors \cite{yin2024sg-nav, hughes2022hydra, xie2024embodiedRAG}. Moreover, the absence of temporal dynamics modeling leaves agents highly vulnerable to obsolete information \cite{nlmap, gu2024conceptgraphs, liu2025delta, tang2025openin}. These limitations restrict the effective resolution of memory discrepancies, which fundamentally impairs the agent's adaptability in dynamic environments.

Furthermore, existing methods relying on exhaustive global search or single-modality information are fundamentally inadequate for object relocation in dynamic environments \cite{liu2025dynamem, yan2025dynamic, park2023generative, zhang2025ella}. Inspired by human cognitive search behaviors, as shown in Fig.~\ref{fig:intention}, individuals locate misplaced objects by leveraging multi-modal contextual cues rather than resorting to random or exhaustive environmental scanning. This process involves synthesizing physical spatial dynamics, such as the influence of recent nearby human interactions on object placement, together with semantic relevance. For example, it recognizes that a missing bowl from a dining table is far more likely to be relocated to a kitchen sink or refrigerator rather than to a semantically unrelated bathroom. Therefore, integrating multi-modal contextual knowledge that encompasses both semantic commonsense and spatio-temporal dynamics into the memory retrieval loop is essential for achieving efficient dynamic object relocation.

To address these challenges, we introduce HitMem, a hierarchical temporal 3D memory framework with multi-modal context-aware retrieval for dynamic environments. HitMem unifies semantic, spatial, and temporal information within a multi-level representation. At the high level, it constructs a lightweight semantic topological graph that encodes only categorical carrier-carried relationships. This bypasses the computationally expensive instance-level correspondence inference and geometric clustering mandated by conventional methods. At the low level, HitMem preserves the detailed 3D attributes of objects, ensuring precise spatial matching and supporting downstream 3D manipulation tasks. To capture the inherently dynamic nature of the environment, we introduce a temporal decay mechanism. It dynamically modulates the activeness scores of memory nodes over time, proactively penalizing outdated observations and preventing the agent from relying on obsolete information during task execution.

Crucially, HitMem incorporates a multi-modal context-aware retrieval mechanism. By sequentially leveraging the semantic and spatial information of the hierarchical memory graph with temporal activeness scores, the system can efficiently filter out invalid targets. When a target is determined to be displaced, we activate a two-stage retrieval strategy that tightly integrates multi-modal environmental context. First, we extract spatial constraints from the historical trajectories of external agents, significantly pruning the search space. Second, we conduct a class affinity analysis that evaluates the probabilistic correlation between the target and candidate carriers. By combining physical trajectory cues with semantic common sense, HitMem avoids blind global exploration and achieves highly targeted, context-driven spatial reasoning.

Considering that existing benchmarks fail to adequately capture the dynamic challenges of real-world environments, we construct a novel benchmark, Dyna-THOR, built upon the AI2-THOR \cite{ai2thor} simulator. Dyna-THOR introduces external agents that actively manipulate and relocate task-relevant objects during execution, simulating the unexpected interactions caused by humans or other agents in real-world settings. Extensive experiments on Dyna-THOR demonstrate that HitMem significantly improves long-term task execution, showing strong adaptability to dynamic disturbances while substantially reducing exploration costs compared to existing baselines.

In summary, our main contributions are as follows:
\begin{itemize}
    \item We propose HitMem, a hierarchical temporal 3D memory that structurally unifies semantic, spatial, and fine-grained geometric information. A temporal decay mechanism dynamically regulates memory activeness, enabling robust adaptation to environmental changes and filtering of obsolete observations.
    \item We design a multi-modal context-aware retrieval mechanism. By integrating physical spatial cues (external agent trajectories) with semantic common sense (class affinity analysis) in a two-stage strategy, the proposed method efficiently and accurately relocates displaced objects.
    \item We introduce Dyna-THOR, a novel benchmark that incorporates interactions between external agents and objects to simulate unpredictable environmental dynamics. Extensive experiments show that HitMem improves relocation accuracy, exploration efficiency, and adaptability in dynamic environments.
\end{itemize}

\section{Related Work}
\label{sec:related}

\subsubsection{3D Scene Representations.}
3D scene representation is crucial for the task execution of embodied agents. Early dense 3D reconstruction approaches \cite{kerr2023lerf, zheng2024gaussiangrasper} capture detailed geometric structures but lack high-level semantics, limiting their effectiveness in tasks involving semantic queries. While integrating Vision-Language Models (VLMs) \cite{clip, li2022blip} embeds 2D semantic features into 3D structures for language-guided object localization and manipulation \cite{mazur2023feature, huang2023voxposer, zhang2023clipfo3d, liu2025dynamem}, these methods remain computationally expensive and incapable of real-time updates in dynamic environments. 3D Scene Graph-based methods further simplify scene representation \cite{lingelbach2023tasksg, yin2024sg-nav, liu2025delta, yan2025dynamic}, offering stronger semantic interpretability through nodes and edges. Accurate scene characterization requires precise edge correspondences, typically achieved through geometric clustering \cite{hughes2022hydra, maggio2024clio} or inference via Large Language Models (LLMs) \cite{kim2024llm4sgg, yin2024sg-nav, xie2024embodiedRAG}. However, these methodologies are structurally unstable in dynamic settings, where object displacements necessitate error-prone and computationally prohibitive recomputations of clusters and edges. Moreover, lacking temporal dynamics modeling leaves agents reliant on obsolete information, degrading task accuracy and efficiency.
In this work, we streamline the 3D scene representation by focusing on categorical carrier-carried relationships. This topological choice bypasses complex edge inference to enable continuous memory updates while preserving fine-grained 3D attributes. Coupled with a temporal decay mechanism, our approach substantially improves agent adaptability and execution efficiency in dynamic environments.

\subsubsection{Memory for Embodied Tasks.}
Embodied agent memory takes various forms. Short-term memory approaches typically employ VLMs for scene analysis and action prediction based on instantaneous observations \cite{brohan2022rt, zitkovich2023rt2, zhou2024navgpt, uni-navid, zhi2025closed}. The absence of a stable and persistent knowledge representation makes it difficult to ensure reliable long-term task execution. Some methods store historical images as long-term memory for embodied reasoning and exploration \cite{cui2024frontier, yang20253d, liu2025spatialcot}, but they often suffer from low efficiency and poor adaptability to dynamic environments. A common approach is to leverage 3D scene representations by directly converting extracted semantic and spatial information into agent memory \cite{nlmap, gu2024conceptgraphs, zhang2025mem2ego, kim2023context}. When needed, the memory can be retrieved to support LLM-based task understanding and planning. Such methods significantly enhance environmental comprehension but are strictly limited to static applications. To address dynamics, recent studies incorporate updates \cite{sun2024dadu, liu2025dynamem, yan2025dynamic} or temporal awareness \cite{park2023generative, zhang2025ella}. However, for object relocation, these methods still resort to global matching or scoring. By relying on exhaustive search and failing to fully exploit environmental context, these approaches suffer from severe inefficiencies in memory retrieval. 
In contrast, we propose a multi-modal context-aware retrieval mechanism leveraging our hierarchical memory structure with temporal dynamics for rapid target localization. To relocate displaced objects, we design a two-stage retrieval strategy that integrates external agent trajectories and semantic class-affinity scanning, significantly reducing the search space and improving retrieval efficiency.

\section{Method}
\label{sec:method}

\begin{figure} [tb]
    \centering
    \includegraphics[width=\linewidth]{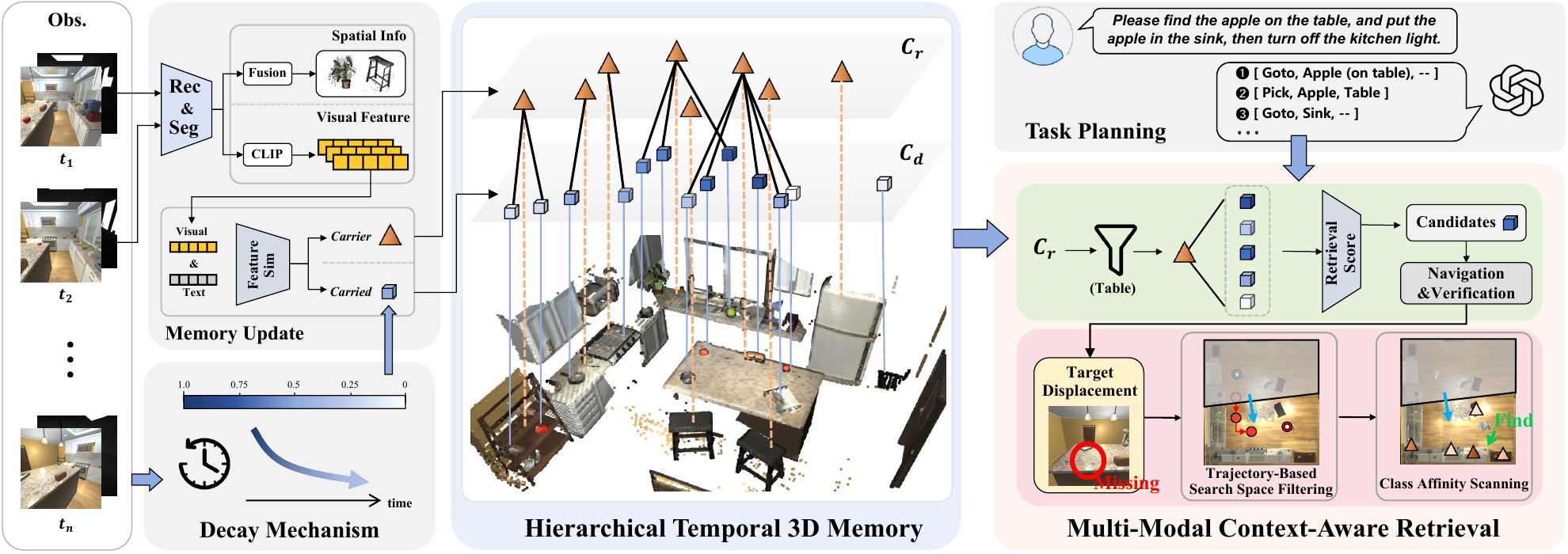}
    \caption{\textbf{Overview of HitMem.} HitMem is a hierarchical 3D memory framework that integrates spatial-semantic information with a decay mechanism to model temporal dynamics. It employs multi-modal context-aware retrieval for efficient target localization. When objects are displaced, it leverages external agent trajectories with class affinity to rapidly identify candidate regions, enhancing robustness in dynamic environments.}
    \label{fig:overview}
\end{figure}

Our objective is to enable embodied agents to execute instructions via continuous RGB-D observations while robustly handling unexpected object displacements in dynamic environments. 
Fig. \ref{fig:overview} illustrates an overview of the proposed HitMem framework. We construct a hierarchical 3D representation that fuses spatial and semantic information, integrated with a decay mechanism to model temporal dynamics. To efficiently locate targets and mitigate memory inconsistencies, HitMem employs a multi-modal context-aware retrieval mechanism. When a target is displaced, it activates a two-stage strategy that synthesizes spatial cues (external agent trajectories) and semantic common sense (class affinities) to rapidly pinpoint candidate regions, ensuring adaptability and robustness.

\subsection{Hierarchical Temporal 3D Memory}
\label{sec:memory}

\subsubsection{Topological and Semantic Representations.}
By continuously perceiving the environment, our approach constructs and dynamically updates an object-centered hierarchical memory graph, denoted as $\mathcal{G}_t = (\mathcal{V}_t, \mathcal{E}_t)$.

\textbf{Node Instantiation ($\mathcal{V}_t$):} Given a continuous stream of observations $\mathcal{I} = \{ I_1, I_2, \dots \}$ from the ego-agent, where each $I_n$ comprises an RGB-D image and the corresponding camera pose, we leverage an open-vocabulary recognition and detection method (e.g., RAM-Grounded-SAM2 \cite{ram, grounding_dino, sam}) to obtain instance-level bounding boxes and segmentation masks. These extractions form the vertex set $\mathcal{V}_t$, with each node $v_i \in \mathcal{V}_t$ encapsulating both semantic and spatial attributes. Specifically, we derive visual embeddings $F_{i}$ by processing cropped images and masks through CLIP \cite{clip}. Concurrently, we associate each node with precise 3D geometric attributes, including the reconstructed point cloud $pcd_i$, global coordinates $pos_i$, and physical dimensions.

\textbf{Carrier-Carried Edges ($\mathcal{E}_t$):} To constrain the search space for downstream tasks, we organize nodes into a hierarchical structure by classifying them into carriers $\mathcal{C}_r$ (e.g., tables, shelves) and carried objects $\mathcal{C}_d$ (e.g., cups, apples). A node $v_i$ is assigned to $\mathcal{C}_r$ if the cosine similarity between its visual feature $F_i$ and the textual embedding of ``furniture or appliances for holding objects'' exceeds a semantic threshold $\theta_{sem}$. For remaining non-carrier nodes $v_j$, we evaluate their spatial relationship with all carriers $v_i \in \mathcal{C}_r$ via a comprehensive support score:
\begin{equation}
    S_{support}(v_i, v_j) = \omega_1 R_{p} + \omega_2 R_{c} + \omega_3 R_{v} + \omega_4 S_{h}
\end{equation}
where $\sum_{k=1}^{4} \omega_k = 1$. This aggregate score encapsulates four geometric metrics derived from the instance-level point clouds: spatial proximity ($R_{p}$), horizontal coverage ($R_{c}$), vertical support ($R_{v}$), and height consistency ($S_{h}$). A directed edge $E(v_i, v_j) \in \mathcal{E}_t$ is established, and $v_j$ is formally assigned to the carried set $\mathcal{C}_d$, only if the maximum support score exceeds a predefined threshold $\theta_{sup}$:
\begin{equation}
    v_j \to \mathcal{C}_d \cup \{ E(\hat{v}_i, v_j) \to \mathcal{E}_t \}, \text{if } \max\limits_{v_i \in \mathcal{C}_r} S_{support}(v_i, v_j) > \theta_{sup}
\end{equation}
If $v_j$ fails to satisfy the support condition for any existing carrier, it is temporarily categorized as an unassigned object ($v_j \in \mathcal{C}_{u}$). These nodes are retained in memory and dynamically re-evaluated for support relationships whenever new carrier nodes are incrementally discovered during subsequent exploration. By abstracting the environment into this dynamic $\mathcal{C}_r \rightarrow \mathcal{C}_d$ topological structure, updates can be executed efficiently via localized node insertion or deletion, circumventing the computational burden of global graph reconstruction.

\subsubsection{Temporal Decay Mechanism.}
\label{sec:decay}

A fundamental challenge in dynamic environments is memory obsolescence, where over-reliance on stale information inevitably leads to memory-observation conflicts and task failures.
To explicitly quantify temporal validity, we introduce a memory decay mechanism that assigns a dynamic activeness score $\mathcal{A}(v_i, t) \in (0,1]$ to each node $v_i \in \mathcal{V}_t$. This score decays exponentially with the time elapsed since the node's last observation:
\begin{equation}
    \mathcal{A}(v_i, t) = \exp \left( - \lambda(v_i) \cdot (t - T_{last}) \right)
\end{equation}

Crucially, instead of using a uniform static constant, we define the decay rate $\lambda(v_i)$ as an intrinsic property of the object's structural role, modulated by the environment's spatial scale. Since carrier objects ($\mathcal{C}_r$) are relatively stationary, their decay rate is set to zero to ensure stable structural memory. Conversely, carried objects ($\mathcal{C}_d$) are assigned a baseline rate $\lambda_{base}$, which is scaled by a scene factor $\eta(S_{scene})$ that grows with the mapped area or room count. Specifically, $\lambda(v_i) = \frac{\lambda_{base}}{\eta(S_{scene})}$. This inverse scaling ensures slower decay in larger environments, preventing premature forgetting during long-term tasks. As a result, a high activeness score indicates reliable memory, while a lower score reduces retrieval priority, encouraging the agent to discount outdated observations.

In real-world scenarios, environments are often perturbed by external agents, such as humans or other autonomous entities, whose interactions alter spatial configurations and introduce unpredictable dynamics. Since these agents provide no explicit notification of displacement, the ego-agent must infer changes solely through observation. Our decay mechanism efficiently supports the tracking of these dynamics. Entities identified as external agents (labeled ``human/robot'') by the perception module are independently recorded as a node set $\mathcal{V}^{agent}$. By ordering these nodes chronologically according to their activeness scores, we reconstruct the external agent's motion to generate a coherent trajectory:
\begin{equation}
    \mathcal{T} = \left( v^{agent}_{k_1}, v^{agent}_{k_2}, \dots, v^{agent}_{k_m} \right), \quad \text{s.t.} \quad \mathcal{A}(v^{agent}_{k_j}) < \mathcal{A}(v^{agent}_{k_{j+1}})
\end{equation}
Visual instance tracking can be integrated to assign identities, allowing the system to partition external agent nodes into subsets and construct independent trajectories for multiple external agents. These trajectories serve as critical physical contexts, guiding multi-modal memory retrieval and target relocation.

\subsubsection{Continuous Memory Fusion.}
To ensure a consistent and deduplicated scene representation, incoming observations must be tightly fused with existing nodes. When integrating new observations, conventional Iterative Closest Point (ICP) matching is prone to segmentation inaccuracies and localization noise, which often absorb small objects into larger ones, leading to the loss of memory entries. To address this, we introduce a Bidirectional ICP matching process requiring both forward and backward alignment scores to pass thresholds ($S_{ICP}^{fwd} \ge \sigma_{fwd} \land S_{ICP}^{bwd} \le \sigma_{bwd}$). Following a successful match, the new point cloud is fused into the stored node $v_i$, its temporal state is refreshed ($\mathcal{A}(v_i, t) = 1$), and its visual feature is updated via a weighted average based on the observation count.

\subsection{Multi-Modal Context-aware Retrieval}
\label{sec:retrieval}

\begin{figure} [tb]
    \centering
    \includegraphics[width=0.95\linewidth]{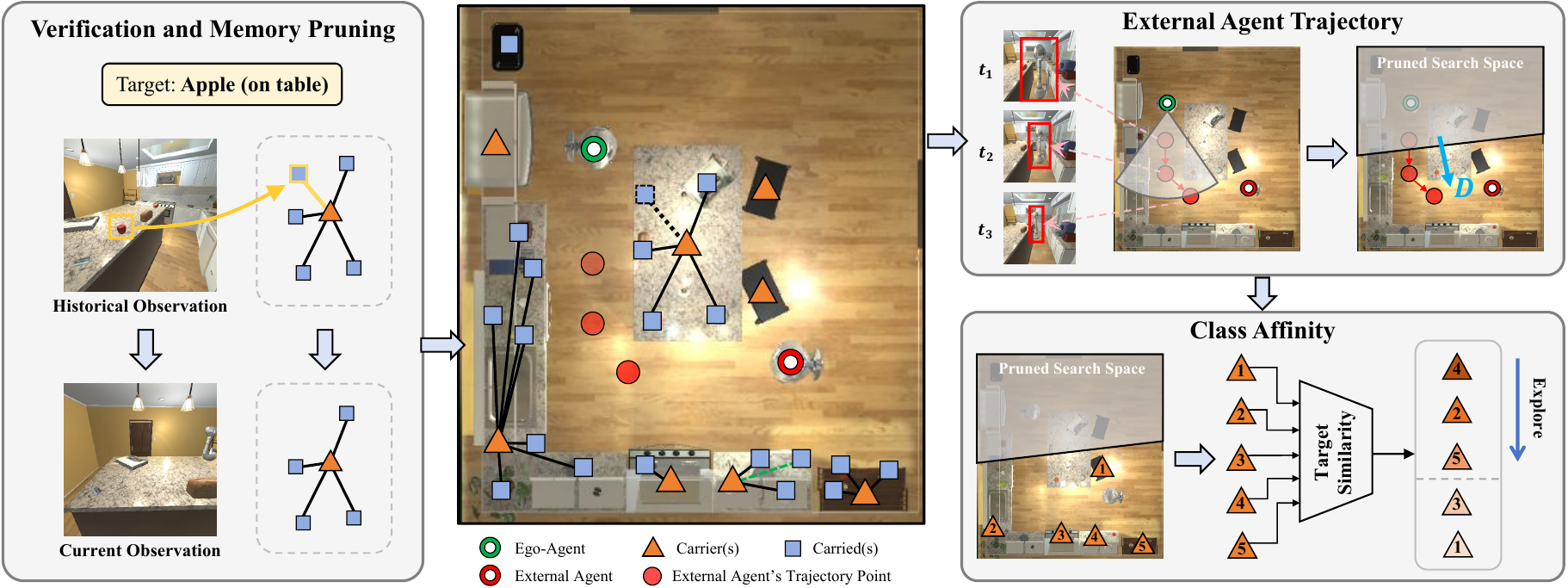}
    \caption{\textbf{Retrieval strategy of HitMem.} Upon navigating to the retrieved target object location, the ego-agent actively verifies its visual observation. If the target has been displaced, the corresponding obsolete memory node is removed, and a two-stage retrieval process is then initiated. First, the local trajectories of external agents are analyzed to infer potential displacement directions and narrow the search space. Second, class affinity scores between the target and candidate carriers are calculated to prioritize subsequent exploration.}
    \label{fig:retrieval}
\end{figure}

Efficient memory retrieval is fundamental for executing high-level task plans in dynamic environments. To address the challenge of target object displacement, we introduce a closed-loop, multi-modal context-aware retrieval method, as shown in Fig.~\ref{fig:retrieval}. The system initially performs a hierarchical semantic query within the memory graph $\mathcal{G}_t$, integrating temporal information to filter candidate targets before conducting active visual verification. Upon detecting a memory-observation conflict signaling object displacement, the system proactively prunes the obsolete node. It then activates a two-stage relocation strategy, synthesizing external agent trajectory and class affinity to perform active reasoning.

\subsubsection{Hierarchical Semantic Querying and Active Verification.}
Leveraging the natural language understanding capabilities of LLMs, user instructions are parsed into structured semantic queries (e.g., $\langle \text{Pick}, \text{Apple}, \text{Table} \rangle$). To minimize computational overhead, HitMem capitalizes on the hierarchical structure of the memory graph $\mathcal{G}_t$. The system initially filters relevant carrier nodes within $\mathcal{C}_r$. Subsequently, it restricts the fine-grained search to the carried objects $v_i \in \mathcal{C}_d$ that are topologically linked to the filtered carriers. In scenarios where the instruction lacks a specific carrier context, the hierarchical topology still accelerates the process by enabling a top-down structured traversal.

We define a comprehensive retrieval score for each candidate carried object: $S_{retrieval}(v_i) = \alpha_1 \cdot \text{sim}(F_i, F_{query}) + \alpha_2 \cdot \mathcal{A}(v_i, t)$. Here, $F_{query}$ denotes the textual embedding of the queried target object, and $\text{sim}(\cdot, \cdot)$ computes the cosine similarity between the visual and textual embeddings. Both the similarity metric and the activeness score are normalized to $[0, 1]$ using min-max scaling. In our implementation, $\alpha_1$ and $\alpha_2$ are set to $1$. Candidates satisfying the requirement $S_{retrieval}(v_i) > \theta_{target}$ are ranked in descending order. Up to three top-ranked nodes are then selected as primary exploration targets, effectively prioritizing objects that possess both semantic relevance and temporal recency.

Upon navigating to the expected location of the top-ranked candidate, the ego-agent verifies whether the current visual observation matches the anticipated target memory. If the target is successfully detected, the retrieval concludes and the downstream manipulation task proceeds. Conversely, if the target is visually absent, HitMem identifies a memory-observation conflict indicating displacement and executes active memory pruning to keep the scene representation consistent. The obsolete node $v_i$ and its associated carrier-carried edge are deleted from the graph ($\mathcal{V}_t \leftarrow \mathcal{V}_t \setminus \{v_i\}$). This proactive mechanism promptly purges verified stale information from memory, ensuring sustained alignment between the ego-agent’s internal representation and the evolving physical environment.

\subsubsection{Two-Stage Retrieval for Relocated Objects.}
When the target object is displaced, the ego-agent initiates a two-stage retrieval strategy. This strategy fully leverages multi-modal environmental context to efficiently re-localize the displaced target, transitioning from blind exploration to guided spatial reasoning.

\textbf{Stage 1: Trajectory-Based Search Space Filtering.}
The initial step leverages the historical trajectory $\mathcal{T}$ of external agents to infer the displacement direction of the target. We first identify the interaction point $\mathcal{P}_{closest} \in \mathcal{T}$, defined as the trajectory node spatially closest to the object's last known position. A local trajectory segment, denoted as $\tau_{local}$, is then extracted, consisting of a subsequence of $k$ consecutive points following $\mathcal{P}_{closest}$. To capture the stable motion trend rather than local fluctuations, we estimate the displacement vector $\vec{D}$ as a weighted average of the subsequent trajectory segments:
\begin{equation}
    \vec{D} = \sum_{m=1}^{k} \left[ \mathcal{A}(v^{agent}_{c+m}) - \mathcal{A}(v^{agent}_{c}) \right] \cdot \left[ pos(v^{agent}_{c+m}) - pos(v^{agent}_{c+m-1}) \right]
\end{equation}
where $c$ is the index of $\mathcal{P}_{closest}$ in $\mathcal{T}$, and the weight is computed based on the activeness score difference. Guided by $\vec{D}$, carrier objects within a sector of angle $\theta_R$ are selected to form the candidate exploration set $\mathcal{C}_{cand}$, formally defined as:
\begin{equation}
    \mathcal{C}_{cand} = \left\{ v_i \in \mathcal{C}_r \;\middle|\; \frac{\vec{D} \cdot (pos_i - pos_{last})}{\| \vec{D} \|_2 \cdot \| pos_i - pos_{last} \|_2} \ge \cos\left(\frac{\theta_R}{2}\right) \right\}
\end{equation}
This process bounds the directional exploration, significantly narrowing the physical search space for the subsequent semantic scanning.

\textbf{Stage 2: Semantic Class Affinity Scanning.}
The second stage prioritizes the candidate carriers in $\mathcal{C}_{cand}$ based on class affinities. To quantify the class affinity between the target category $Cat_t$ and each candidate carrier $v_j \in \mathcal{C}_{cand}$, we populate $Cat_t$ into a descriptive text template (e.g., ``A receptacle where a [$Cat_t$] is most often placed'') to extract a contextualized textual embedding $E_{text}(Cat_t)$. The raw semantic affinity $A_{affinity}(Cat_t, v_j)$ is then evaluated by computing the cosine similarity between the textual and visual embeddings of the candidate carrier. Alternatively, this score can be replaced by a commonsense prior inferred by an LLM. To form a valid probability distribution for sequential exploration, these values are normalized across the entire candidate set $\mathcal{C}_{cand}$ using a standard softmax function:
\begin{equation}
    P(v_j \mid Cat_t) = \frac{\exp \left( A_{affinity}(Cat_t, v_j) \right)}{\sum_{v_k \in \mathcal{C}_{cand}} \exp \left( A_{affinity}(Cat_t, v_k) \right)}
\end{equation}
By directing the ego-agent to explore candidate carriers prioritized by $P(v_j \mid Cat_t)$, this mechanism significantly reduces redundant exploration.
\section{Experiments}
\label{sec:exp}
To comprehensively evaluate our HitMem framework, we first introduce Dyna-THOR, a novel benchmark constructed to address the critical lack of interaction-driven environmental dynamics. We conduct comparative experiments to assess the performance of various methods in dynamic and long-term task scenarios. Furthermore, we design ablation studies and extension experiments to validate the effectiveness and scalability of our approach across diverse settings. Qualitative visualizations are provided to intuitively interpret the operational process.

\subsection{Dyna-THOR Design}
\label{sec:dynathor}

\begin{table}[tb]
  \scriptsize
  \setlength{\tabcolsep}{6pt}
  \caption{Comparison of existing embodied AI benchmarks}
  \label{tab:benckmark_comparison}
  \centering
  \begin{tabular}{@{}lcccc@{}}
    \toprule
     & Vis. Obs. & \makecell{Movable \\ Objects}  & \makecell{Long-term \\ Tasks} & \makecell{Dynamism} \\
    \midrule
    R2R \cite{anderson2018R2R} & \textcolor{blue}{Ego} & \textcolor{red}{\ding{55}}  & \textcolor{red}{\ding{55}} & \textcolor{red}{\ding{55}} \\
    VirtualHome \cite{puig2018virtualhome} & \textcolor{red}{$3^{rd}$ Person} & \textcolor{blue}{\ding{55}}  & \textcolor{blue}{\ding{51}} & \textcolor{red}{\ding{55}} \\
    ALFRED \cite{ALFRED20} & \textcolor{blue}{Ego} & \textcolor{blue}{\ding{51}}  & \textcolor{red}{\ding{55}} & \textcolor{red}{\ding{55}} \\
    \textbf{Dyna-THOR (Ours)} & \textcolor{blue}{Ego} & \textcolor{blue}{\ding{51}} & \textcolor{blue}{\ding{51}} & \textcolor{blue}{\ding{51}} \\
    \bottomrule
  \end{tabular}
\end{table}

Existing embodied AI datasets and benchmarks primarily focus on navigation, reasoning, and manipulation tasks within static environments \cite{xiazamirhe2018gibsonenv, anderson2018R2R, puig2018virtualhome, ALFRED20, kim2024realfred}. As shown in Table \ref{tab:benckmark_comparison}, existing benchmarks are typically constructed in non-interactive, static environments or overlook the impact of object manipulations performed by other agents (e.g., humans). Consequently, they fail to effectively simulate the challenges encountered by robots operating in real-world settings, where unanticipated human interactions frequently alter object locations during long-term task execution. This limitation results in insufficient evaluation of a method’s robustness and adaptability under realistic dynamic conditions.

To bridge this critical gap, we introduce Dyna-THOR, a novel benchmark built upon the AI2-THOR simulator \cite{ai2thor}. Its core evaluation mechanism extends beyond merely requiring the \textit{ego-agent} to accomplish predefined tasks. Specifically, we incorporate \textit{external agents} that interact with environmental objects to simulate realistic human behaviors.  These external agents deliberately manipulate task-relevant objects by altering their physical states or relocating them prior to the ego-agent's execution. Crucially, these interventions occur without providing any explicit notification to the ego-agent. By explicitly modeling the root causes of real-world environmental changes, this design intentionally decouples the ground-truth state from the ego-agent’s internal memory, thereby naturally inducing memory-observation conflicts through unpredictable dynamics during task execution, enabling a highly rigorous and precise evaluation of various methods regarding their robustness and adaptability in dynamic settings.

Dyna-THOR comprises 12 interactive scenes across four distinct room categories, each featuring five long-term tasks. In these tasks, the ego-agent must interpret and sequentially execute multiple natural language instructions to achieve specified objectives, while simultaneously adapting to environmental dynamics induced by external agents. More details about the Dyna-THOR dataset are provided in the supplementary materials.

\subsection{Experimental Setup}
\label{sec:setup}

\textbf{Task and Environment Configurations.} 
We conduct experiments in the AI2-THOR simulator. The ego-agent processes egocentric RGB-D observations at a resolution of $400 \times 400$ pixels. For task execution, we enforce a strict maximum limit of 500 steps. If the ego-agent exceeds this threshold, the episode is immediately terminated, and the final environmental state is evaluated against the expected goal. 
The action space of the ego-agent includes navigation actions (MoveAhead-0.25m, RotateLeft-90\textdegree , RotateRight-90\textdegree ), manipulation actions (e.g., PickupObject, PutObject, OpenObject, CloseObject, ...), and a termination action (Done).

\textbf{Baselines.} 
We compare our method with the following baselines: 1) DELTA \cite{liu2025delta}: This method leverages a label-based scene graph for task understanding and planning with LLMs. 2) ConceptGraphs \cite{gu2024conceptgraphs}: It constructs a static open-vocabulary 3D scene graph for object-level semantic mapping, enabling feature-based matching and querying. 3) DovSG \cite{yan2025dynamic}: It maintains a dynamic 3D scene graph that supports localized updates of spatial and semantic relationships. 4) DynaMem \cite{liu2025dynamem}: It proposes an online updatable spatio-temporal semantic point cloud memory, which dynamically reconstructs spatial coordinates in response to environmental changes and supports open-vocabulary object localization.

To ensure a fair comparison, all methods requiring pre-constructed memory are provided with an identical initial exploration trajectory. Every method employs the identical recognition and segmentation models for visual processing and utilizes the GPT-4o for all LLM-based reasoning and planning tasks. All methods are restricted to navigating via memory-retrieved spatial coordinates rather than relying on simulator-defined ground-truth object identifiers.

\textbf{Evaluation Metrics.} 
We quantitatively assess task performance using three primary metrics. \textit{Success Rate (SR):} This metric measures the overall task completion. A trial is deemed successful only if all necessary actions are executed and the final environmental state strictly matches the ground-truth goal state. \textit{Success weighted by Path Length (SPL):} This metric evaluates the physical spatial exploration efficiency by calculating the ratio of the theoretically shortest path to the actual executed path length. \textit{Ground-truth Completion Rate (GCR):} We additionally employ this metric to measure the consistency of task execution across various natural language instructions within each long-term task.

\subsection{Comparison Result}
\subsubsection{Comparison with Baselines.}

\begin{table}[tb]
\centering
\fontsize{4.7}{6}\selectfont
\renewcommand{\arraystretch}{1.4}
\caption{Detailed results on the Dyna-THOR.}
\begin{tabularx}{\textwidth}{@{\hspace{0.2em}} p{1.75cm} *{15}{>{\centering\arraybackslash}X} @{\hspace{0.3em}}}
\toprule
\multirow{2}{*}{\hfil\textbf{Method}} & \multicolumn{3}{c}{\textbf{Kitchen}} & \multicolumn{3}{c}{\textbf{Living-room}} & \multicolumn{3}{c}{\textbf{Bedroom}} & \multicolumn{3}{c}{\textbf{Bathroom}} & \multicolumn{3}{c}{\textbf{Overall}} \\
\cmidrule(lr){2-4} \cmidrule(lr){5-7} \cmidrule(lr){8-10} \cmidrule(lr){11-13} \cmidrule(lr){14-16}
 & \textbf{SR$\uparrow$} & \textbf{SPL$\uparrow$} & \multicolumn{1}{c|}{\textbf{GCR$\uparrow$}} & \textbf{SR$\uparrow$} & \textbf{SPL$\uparrow$} & \multicolumn{1}{c|}{\textbf{GCR$\uparrow$}}
 & \textbf{SR$\uparrow$} & \textbf{SPL$\uparrow$} & \multicolumn{1}{c|}{\textbf{GCR$\uparrow$}} &  \textbf{SR$\uparrow$} & \textbf{SPL$\uparrow$} & \multicolumn{1}{c|}{\textbf{GCR$\uparrow$}} & \textbf{SR$\uparrow$} & \textbf{SPL$\uparrow$} & \textbf{GCR$\uparrow$} \\
\midrule
DELTA \cite{liu2025delta} & 0.00 & 0.00 & \multicolumn{1}{c|}{15.56} & 6.67 & 5.63 & \multicolumn{1}{c|}{22.22} & 0.00 & 0.00 & \multicolumn{1}{c|}{31.89} & 6.67 & 5.24 & \multicolumn{1}{c|}{23.89} 
& 3.33 & 2.72 & 23.39\\

ConceptGraph \cite{gu2024conceptgraphs} & 6.67 & 2.32 & \multicolumn{1}{c|}{23.89} & 13.33 & 8.19 & \multicolumn{1}{c|}{28.89}
& 13.33 & 6.95 & \multicolumn{1}{c|}{36.89} & 13.33 & 10.71 & \multicolumn{1}{c|}{17.78}
& 11.67 & 7.04 & 26.86 \\

DovSG \cite{yan2025dynamic} & 13.33 & 8.03 & \multicolumn{1}{c|}{32.22} & 13.33 & 10.94 & \multicolumn{1}{c|}{38.89}
& 26.67 & 18.89 & \multicolumn{1}{c|}{50.22} & 20.00 & 14.50 & \multicolumn{1}{c|}{37.22}
& 18.33 & 13.09 & 39.64 \\

DynaMem \cite{liu2025dynamem} & 33.33 & 14.78 & \multicolumn{1}{c|}{43.33} & 40.00 & 17.29 & \multicolumn{1}{c|}{52.22}
& 40.00 & 17.20 & \multicolumn{1}{c|}{58.00} & 26.67 & 18.16 & \multicolumn{1}{c|}{37.22}
& 35.00 & 16.86 & 47.69 \\

HitMem(Ours) & \textbf{40.00} & \textbf{24.97} & \multicolumn{1}{c|}{\textbf{56.67}} & \textbf{53.33} & \textbf{36.80} & \multicolumn{1}{c|}{\textbf{65.56}}
& \textbf{66.67} & \textbf{46.32} & \multicolumn{1}{c|}{\textbf{73.00}} & \textbf{46.67} & \textbf{28.86} & \multicolumn{1}{c|}{\textbf{61.11}}
& \textbf{51.67} & \textbf{34.24} & \textbf{64.08} \\ 
\bottomrule
\end{tabularx}
\label{tab:exp_dynathor}
\end{table}

Table~\ref{tab:exp_dynathor} presents a detailed comparison on Dyna-THOR. The results show that HitMem significantly outperforms all baselines in both success rate and execution efficiency, demonstrating its strong capability for robust and efficient long-term task execution in dynamic environments. 

Static memory approaches (DELTA and ConceptGraphs) lack effective memory update mechanisms and rely primarily on outdated scene information despite environmental changes. Given that each long-term task in Dyna-THOR involves object displacement introduced by external agents, these methods struggle to adapt, leading to low success rates. Their few successful cases occur only when object displacement is minor and the moved object remains visible from its original memorized position. ConceptGraph achieves higher performance than DELTA by utilizing feature-based matching. This suggests that feature-based methods offer greater robustness against misclassifications arising from recognition models compared to label-based methods.

Although DovSG supports dynamic updates of the 3D scene graph, its update strategy is limited to local observations. Specifically, memory is refreshed only before manipulation actions, without continuous updating during navigation and execution. As a result, it can handle only object displacements that occur near their original locations. While this design offers some improvement over fully static methods, the performance gain remains limited. DynaMem supports continuous dynamic memory updates and can trigger re-exploration when a target is lost. However, it is constrained by the inefficiency of its global search strategy. This exhaustive exploration often causes the agent to exceed the maximum step limit during the relocation phase. In contrast, HitMem leverages multi-modal context-aware retrieval to drastically narrow the search space and locate targets rapidly. Consequently, our method exhibits superior performance in both overall task completion and execution efficiency.

\subsubsection{Ablation Study.}

Table~\ref{tab:exp_ablation} quantifies the individual contributions of HitMem's core components. Removing the hierarchical structure ($M_H$) noticeably degrades success rate and navigation efficiency. The topology provides essential high-level semantics for task goal understanding. Without this semantic stratification, the agent cannot effectively filter the search space using relevant carrier nodes, which increases exploration path length and impairs re-localization of displaced objects. Disabling the temporal decay mechanism ($M_D$) results in a severe drop in the SPL metric. Without activeness scoring, the agent treats stale and fresh observations equally, leading to repeated navigation to outdated coordinates. This results in wasted steps on invalid targets, such as ``ghost entries'' of objects displaced by external agents. The most substantial degradation occurs when both the context-aware retrieval strategy and $M_D$ are removed. Our full strategy leverages external agent trajectories and semantic affinity to predict probable locations, effectively simulating human reasoning. Conversely, lacking this guidance forces the agent into blind global searches. Such ineffective wandering exhausts the step limit, critically compromising overall task execution.

\begin{table} [tb]
  \caption{Ablation studies on different components.}
  \label{tab:exp_ablation}
  \centering
  \begin{tabular}{ @{\hspace{6pt}} p{5cm} >{\centering\arraybackslash}p{1.4cm} >{\centering\arraybackslash}p{1.4cm} >{\centering\arraybackslash}p{1.4cm} @{\hspace{6pt}}}
    \toprule
      \textbf{Method} & \textbf{SR} $\uparrow$ & \textbf{SPL} $\uparrow$ & \textbf{GCR} $\uparrow$\\
    \midrule
    w/o Hierachical Structure ($M_H$) & 46.67 & 26.71 & 61.43  \\
    w/o Decay Mechanism ($M_D$) & 43.33 & 20.45 & 52.97 \\
    w/o Retrieval Strategy \& $M_D$ & 31.67 & 12.66 & 42.97 \\
    HitMem (Ours) & \textbf{51.67} & \textbf{34.24} & \textbf{64.08} \\
    \bottomrule
  \end{tabular}
\end{table}

\begin{figure} [tb]
    \centering
    \includegraphics[width=0.5\linewidth]{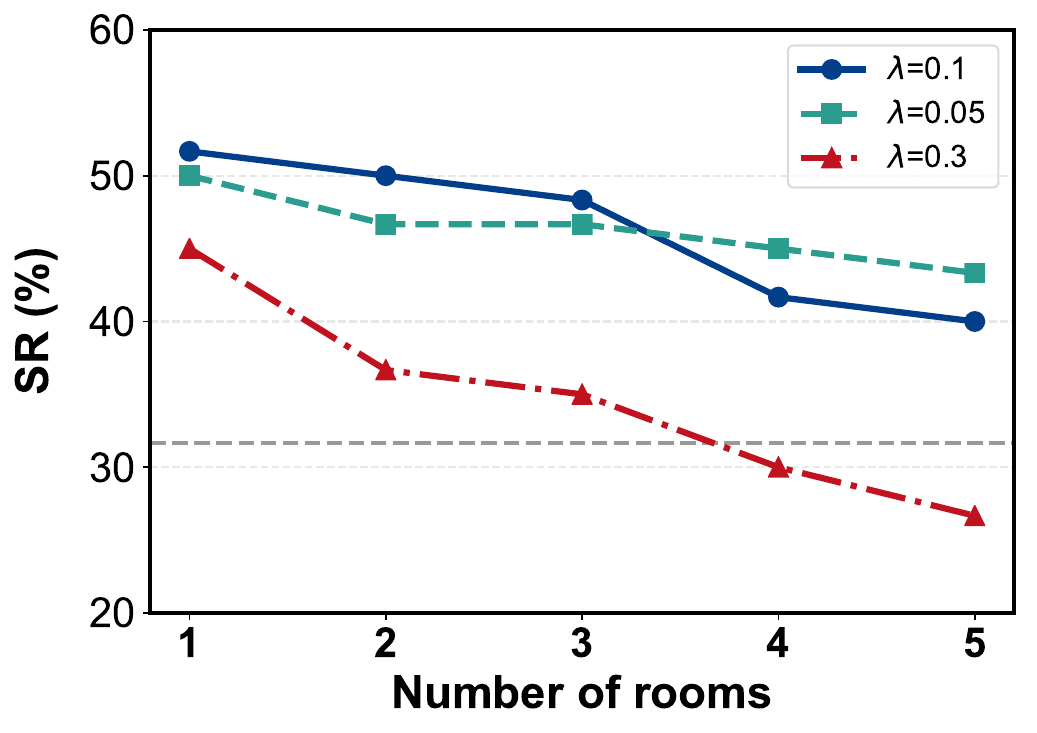}
    \caption{Performance across different scene scales and decay rate settings.}
    \label{fig:exp_scalability}
\end{figure}

\subsection{Scalability Analysis and Visualizations}

\subsubsection{Scalability Analysis.} We further utilize the multi-room environments provided by RoboTHOR \cite{deitke2020robothor} to evaluate the robustness of HitMem across varying spatial scales and to analyze the impact of the temporal decay rate ($\lambda$) on overall performance. Fig.~\ref{fig:exp_scalability} illustrates the success rate trends under different room quantities and decay configurations. 

In small-scale environments, performance variations across different decay rates remain relatively minor, as frequent observations enable timely memory updates. However, the choice of $\lambda$ becomes critical in larger spatial configurations. Values of $\lambda$ that are optimal in single-room scenarios may yield performance degradation when applied to multi-room settings, where reduced revisit frequencies for the same object cause historical memories to continuously decay, thereby lowering their retrieval priority. Adopting a smaller $\lambda$ can alleviate this issue by enhancing memory persistence. Conversely, an excessively large $\lambda$ leads the agent to over-prioritize recent observations, resulting in narrowly-focused searches that fail to track substantially displaced targets and consequently cause a sharp decline in success rates. Therefore, dynamically adjusting the decay rate according to the spatial scale can effectively balance the persistence of memory with the adaptability required for dynamic environments.

Under identical decay settings, expanding the exploration space from a single-room to multi-room environments naturally increases both maintenance and exploration costs. This expanded search space leads to an anticipated downward trend in the overall success rate. To evaluate robustness, we establish the optimal success rate of the variant lacking a decay mechanism in the single-room scenario as a fixed baseline (gray dashed line). With an appropriate decay rate, our approach consistently surpasses the fixed baseline. This confirms that HitMem, powered by hierarchical temporal memory and multi-modal context-aware retrieval, reliably boosts task performance across varying spatial scales.

\begin{figure} [tb]
    \centering
    \includegraphics[width=\linewidth]{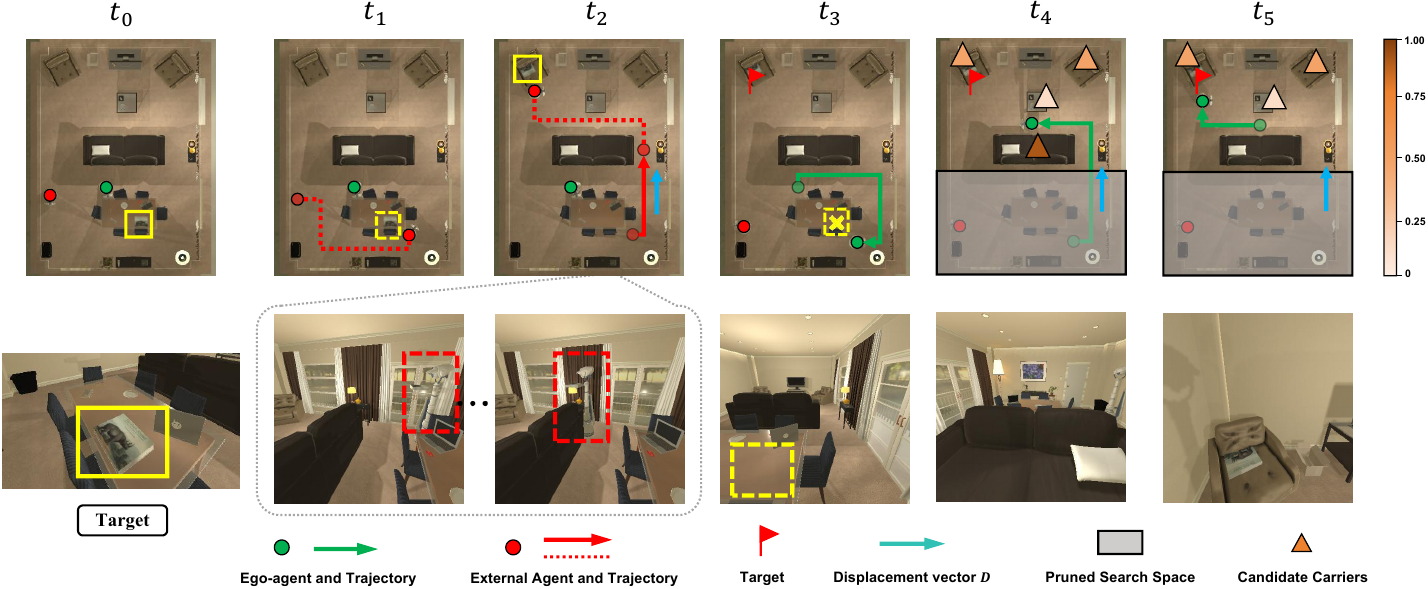}
    \caption{Visualization of a representative dynamic object relocation process.}
    \label{fig:visualization}
\end{figure}

\subsubsection{Case Visualization.}

Fig.~\ref{fig:visualization} provides a visualization of a dynamic object relocation process. It illustrates the movement trajectories of both the ego-agent and the external agent, along with the ego-agent’s active visual observations. At $t_1$, the external agent removes the task relevant target object (Book) and places it in a new location. Between $t_1$ and $t_2$, the ego-agent partially observes and records the movement sequence of the external agent. Upon receiving the instruction to find the book, the ego-agent first queries its internal memory to retrieve the last known coordinates of the target. Arriving at this expected location at $t_3$, the ego-agent experiences a conflict between its memory and current observations because the book is visually absent. To recover from this failure, the ego-agent infers a potential displacement vector utilizing the recorded trajectory points of the external agent. This geometric operation effectively filters the global search space. Subsequently, the system calculates a prioritized set of candidate carriers based on semantic class affinity. Guided by this ranked list, the agent navigates to the most probable carrier (a sofa) at $t_4$. After a local visual inspection yields no results, the agent continues its sequential exploration. Ultimately at $t_5$, the agent successfully detects the target object on the next highest ranked carrier (an armchair) and proceeds with the remaining task execution.
\section{Conclusion}
\label{sec:conclusion}

In this paper, we propose HitMem, a novel hierarchical temporal memory framework designed for embodied agents performing long-term tasks in dynamic environments. HitMem unifies semantic, spatial, and temporal information into a lightweight topological graph while employing a temporal decay mechanism to dynamically modulate the activeness scores of memory nodes. Based on this memory representation, we further introduce a multi-modal context-aware retrieval mechanism for efficient target object localization. We specifically design a two-stage retrieval strategy that combines physical trajectory cues with semantic class affinity to reduce redundant exploration and improve search efficiency. 
Furthermore, we construct the Dyna-THOR benchmark to explicitly simulate unpredictable interferences from external agents. Extensive evaluations demonstrate that HitMem effectively adapts to dynamic environmental changes and scales well across different scenarios, while significantly reducing exploration costs during complex task execution.

\section*{Acknowledgements}
This work is supported by the National Natural Science Foundation of China 62472412 and Project of Institute
of Software, Chinese Academy of Sciences (ISCAS-JCMS-202402).

%
%
\bibliographystyle{splncs04}
\bibliography{main}

\clearpage
\appendix
\renewcommand{\theHsection}{appendix.\Alph{section}}
\section*{\centering Appendix}

\section{Dyna-THOR Benchmark}

\subsection{Action Space and Environment Configuration}

The Dyna-THOR benchmark is built upon the widely used AI2-THOR interactive simulation platform. To focus the evaluation on high-level planning and long-term memory, we abstract the primitive low-level action space by introducing navigation-centric action primitives, specifically \textit{GoToObject} and \textit{GoToPosition}. By encapsulating atomic \textit{Move} and \textit{Rotate} commands, this abstraction supports smooth, continuous navigation via high-level interfaces, thereby allowing for a more accurate assessment of approaches to scene understanding, memory construction, and task planning. Regarding implementation, rotation is discretized into fixed 90-degree increments (\textit{RotateLeft90} and \textit{RotateRight90}). The \textit{GoTo} primitives employ the A* algorithm to compute the shortest path. To ensure valid transitions within the grid-based environment, where manipulation tasks may change the agent's facing direction, we perform a pre-navigation check that calibrates the agent's orientation and ensures it remains orthogonally aligned with reachable grid points. Furthermore, we incorporate a collision avoidance mechanism. During path planning, the algorithm explicitly excludes grid points obstructed by static object colliders or dynamically occupied by other agents. The deterministic nature of the algorithm ensures that navigation trajectories are strictly reproducible under identical environmental configurations, effectively eliminating stochastic variables. For other manipulation actions, we retain the standard definitions provided by the simulator.

In the experimental setup, we initialize two agents within the environment: an \textit{Ego-Agent} and an \textit{External Agent}. While both agents possess identical capabilities for object manipulation, they differ in their navigation privileges. 
The external agent is endowed with the \textit{GoToObject} privilege, enabling it to navigate using ground-truth information, such as object IDs. We leverage this capability to precisely localize and manipulate task-relevant objects, thereby simulating the environmental dynamics introduced by human activities in the real world through the external agent's behaviors.
In contrast, the ego-agent is restricted to \textit{GoToPosition} and is prohibited from accessing environmental metadata for guidance. As a result, it is compelled to depend entirely on its internal memory representations to determine target coordinates. By implementing the aforementioned environmental configuration, we standardize the experimental context, mitigating interference arising from environmental variability. 
This design isolates the method's core evaluation, enabling a precise quantification of how the agent's memory and planning capabilities contribute to overall task performance.

\subsection{Benchmark Composition}

\begin{figure*} [tb]
    \centering
    \includegraphics[width=\linewidth]{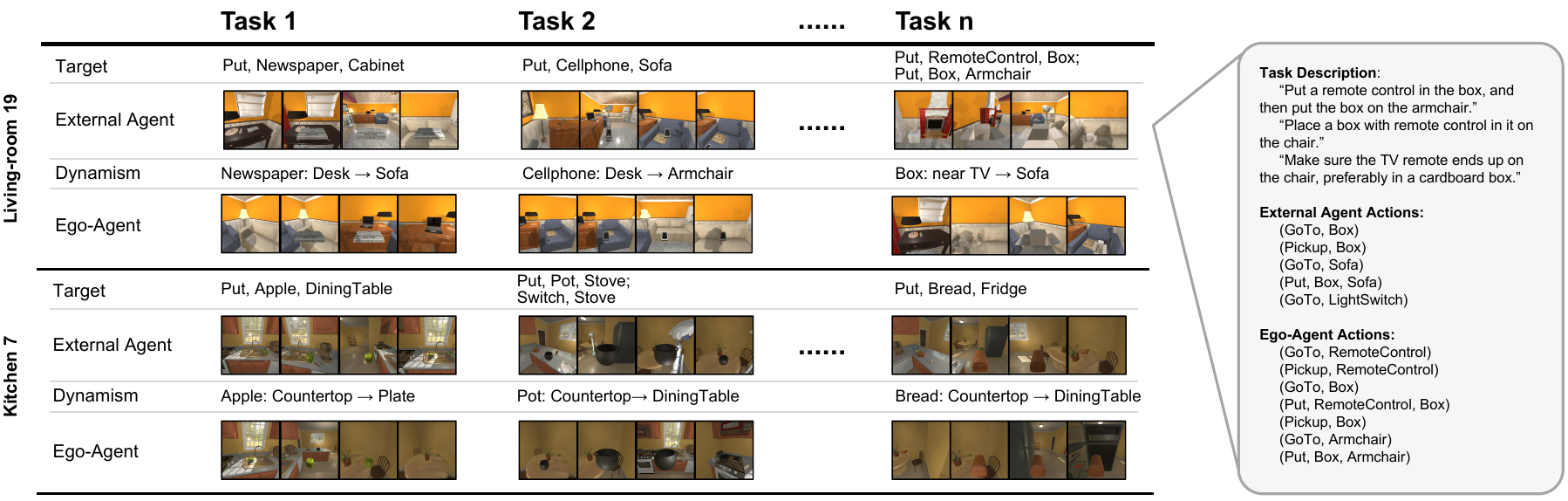}
    \caption{\textbf{Examples from the Dyna-THOR benchmark.} We curated 12 distinct scenes spanning the four categories defined by AI2-THOR, encompassing a total of 60 long-term tasks. Each task is characterized by a specific set of definitions, external agent actions, and ego-agent actions. The figure above illustrates representative episodes derived from expert demonstrations.}
    \label{fig:sup_benchmark}
\end{figure*}

Each data sample constitutes a complete task episode, encapsulating comprehensive metadata to support rigorous evaluation. A standard episode comprises the following core components:

\textbf{Agent Initialization and Configuration.} To ensure reproducibility, each episode explicitly defines the initial states for both the ego-agent and the external agent. It includes precise 3D coordinates, rotational orientation, camera horizon, and posture. This standardized initialization guarantees that both agents operate within a deterministic context before dynamic interactions commence.

\textbf{Natural Language Task Instructions.} Each long-term task involves multiple sub-goals driven by high-level natural language instructions. To evaluate the agent's semantic understanding and robustness, each task intent is paired with distinct linguistic variations. For instance, a retrieval task might be phrased explicitly as ``Go to the book and pick it up,'' or contextualized as ``I left my book in the kitchen, please help me find it.'' This linguistic diversity serves as a rigorous benchmark to assess how different methods and models generalize across varying phrasings, distinguishing true semantic comprehension from the mere memorization of command templates.

\textbf{Agent Action Sequences.} 
Unlike existing static benchmarks, Dyna-THOR introduces dynamic manipulations (external agent actions) to mimic human behavior in real-world environments. These sequences are designed to deliberately manipulate objects relevant to the current task goal before the ego-agent begins execution, thereby introducing environmental dynamics that adhere to natural physical laws. The dataset explicitly details the specific operations performed by the external agent for each task, such as moving a target book from its original location to a sofa. Such actions break the static environment assumption and compel the ego-agent to update its internal memory and replan its actions to adapt to these changes.
In parallel, expert trajectories (ego-agent actions) represent the optimal understanding and execution of the task. These trajectories serve as ground truth action sequences required for efficient task completion and are produced by directly invoking the \textit{GoToObject} command with specific object category IDs within the environment. They embody an accurate task workflow characterized by a minimal and non-redundant sequence of operations. During the actual evaluation of different methods, the ego-agent is denied access to these expert trajectories and must instead rely on its own integrated approach for real-time planning and execution.

\textbf{Goal States and Metrics.} Each episode defines the ground-truth terminal states of key objects upon task completion and records the total step count required by the ego-agent based on the expert trajectory. This step count is quantified by the number of low-level environment primitives executed. These annotations provide a basis for calculating success rates and benchmarking the actual efficiency of long-term planning algorithms.

\subsection{Expert Demonstrations}
The task definition specifies an optimal execution sequence, representing the minimal set of high-level actions required for the ego-agent to achieve its goals. Constrained by environmental interaction rules, the agent must navigate to the immediate vicinity of a target object before initiating any manipulation. The expert demonstration strictly adheres to this optimal sequence, utilizing \textit{*Object} actions (e.g., \textit{GoToObject}). This mechanism enables the expert to leverage ground-truth object IDs for direct localization, ensuring precise navigation to the target even if it has been displaced, thereby eliminating the need for search or exploration. We define Minimum Execution Steps as the total count of low-level action primitives within this expert trajectory. Fig.~\ref{fig:sup_benchmark} illustrates representative examples derived from these expert demonstrations.

The left panel of Fig.~\ref{fig:sup_benchmark} visualizes the environmental observations for two long-term tasks under expert demonstration. Each task comprises multiple sub-goals that must be resolved sequentially. Throughout this process, the ego-agent is required to adapt to positional shifts of target objects caused by external agent interference. A long-term task is deemed complete only upon the successful execution of all constituent sub-goals.
The right panel details the composition of a specific instance: \textit{Livingroom 19-Task n}. First, it presents natural language task descriptions with varied phrasing, employing diverse levels of descriptive granularity to rigorously evaluate the method's capabilities in semantic understanding and planning. Second, it defines the action sequence for the external agent, which injects dynamism into the environment to simulate the uncertainty inherent in real-world human activities. Finally, it provides the corresponding optimal execution sequence for the ego-agent.
Taking this task as a concrete example, the objective is to locate a remote, place it into a box, and subsequently position the box on an armchair. However, the benchmark-introduced dynamics intervene prior to the ego-agent's execution: the external agent relocates the box from its initial position near the TV to a sofa. Consequently, the ego-agent fails to find the target at the expected location and must replan its strategy. It is forced to explore the environment to determine the box's new position, a critical step required to ensure the successful completion of subsequent sub-goals.

\section{Additional Experiment Results}

\subsection{Impact of Descriptions and LLMs}

Table~\ref{tab:sup_exp_ablation_3} presents the experimental results obtained by varying the input task descriptions and utilizing different Large Language Models (LLMs) for task planning. We observe that the granularity of natural language descriptions significantly impacts task success rates. When instructions are explicit and intuitive, LLMs can readily comprehend task requirements and generate correct action sequences. Conversely, the use of ambiguous descriptions leads to a marked decline in success rates. This is because ambiguity tends to obscure critical task information, disrupting the LLM's reasoning process and resulting in the generation of erroneous action sequences or incorrect target identification, which severely compromises task execution.

\begin{table} [htbp]
  \setlength{\tabcolsep}{6pt}
  \caption{Success Rate by using different LLMs and descriptions.}
  \label{tab:sup_exp_ablation_3}
  \centering
  \begin{tabular}{@{\hspace{6pt}}lccc@{\hspace{6pt}}}
    \toprule
      LLM & \textbf{Clear} & \textbf{Medium} & \textbf{Vague}  \\
    \midrule
    Gemini-2.5-pro & 48.33 & 40.00 &  31.67  \\
    GPT-4o & 51.67 & 41.67 & 28.33 \\
    \bottomrule
  \end{tabular}
\end{table}

Simultaneously, we note distinct disparities in task understanding across different models, which can be attributed to variations in their inference capabilities and action generation strategies. Specifically, while Gemini-2.5-Pro possesses robust reasoning capabilities, it tends to generate redundant action sequences. These superfluous actions incur additional exploration costs and elongate trajectories, thereby increasing the risk of exceeding maximum step limits. Consequently, this computational and operational overhead results in a slight degradation in task completion performance.

Therefore, practical deployment necessitates a strategic selection of LLMs tailored to specific task characteristics. Particularly in resource-constrained environments, it is crucial to balance the trade-offs among model capability, planning fidelity, and response latency to achieve optimal performance and efficiency. Concurrently, to further mitigate execution risks, task instructions should be formulated with high precision, explicitly minimizing semantic ambiguity.

\subsection{Resilience to Imperfect Tracking}

We conducted additional ablation studies to evaluate the robustness of our method against various external agent tracking challenges, including frame loss and noise perturbations. The results are shown in Table~\ref{tab:exp_ablation_add}. When simulating ``off-screen'' displacements where tracking fails (affinity-only), the system still achieves a robust 46.67\% SR, outperforming the Trajectory-only setting, which struggles with inefficient physical exploration without semantic guidance. Furthermore, to evaluate robustness against imperfect perception, we introduced trajectory disturbances. Both randomly dropping 30\% of trajectory points (simulating frame loss) and injecting Gaussian noise ($\sigma$=0.02, simulating depth sensor noise) only resulted in a marginal performance drop. These findings prove that our system does not rigidly depend on perfect trajectory tracking, and that the synergy between spatial pruning and semantic commonsense ensures strong resilience against noisy or missing observations.

\begin{table} [htbp]
  \caption{Ablation study on external agent tracking resilience.}
  \label{tab:exp_ablation_add}
  \centering
  \begin{tabular} { @{\hspace{6pt}} p{5.0cm} >{\centering\arraybackslash}p{1.4cm} >{\centering\arraybackslash}p{1.4cm} >{\centering\arraybackslash}p{1.4cm} @{\hspace{6pt}}}
    \toprule
      \textbf{Method} & \textbf{SR} $\uparrow$ & \textbf{SPL} $\uparrow$ & \textbf{GCR} $\uparrow$\\
    \midrule
    Trajectory-only (w/o Affinity) & 40.00 & 20.58 & 49.50  \\
    Affinity-only (Off-screen) & 46.67 & 25.76 & 55.61 \\
    30\% Trajectory Dropout & 50.00 & 30.44 & 60.19 \\
    Noisy Trajectory & 50.00 & 29.45 & 59.64 \\
    Full Method & \textbf{51.67} & \textbf{34.24} & \textbf{64.08} \\
    \bottomrule
  \end{tabular}
\end{table}

\subsection{Generalizability}
We evaluated HitMem on a separate, widely recognized benchmark ALFRED. To evaluate zero-shot generalization under dynamic conditions, we employed the valid\_unseen split (comprising 85 tasks) and introduced artificial dynamic disturbances by relocating the target object to an alternative valid receptacle during ego-agent execution. As shown in the Table~\ref{tab:exp_generalizability}, HitMem outperforms other dynamic memory baselines both on SR and SPL. This demonstrates that our hierarchical temporal memory and multi-modal context-aware retrieval strategy generalize robustly to new domain distributions.

\begin{table}[htbp]
\centering
\caption{Generalizability comparison across different methods.}
\label{tab:exp_generalizability}
\begin{tabular}{ @{\hspace{6pt}} p{3.0cm} >{\centering\arraybackslash}p{1.4cm} >{\centering\arraybackslash}p{1.4cm} @{\hspace{6pt}}}
\toprule
\textbf{Method} & \textbf{SR} $\uparrow$ & \textbf{SPL} $\uparrow$ \\
\midrule
    DovSG & 35.29 & 18.81  \\
    DynaMem & 41.18 & 17.94  \\
    HitMem (Ours) & \textbf{52.94} & \textbf{26.72}  \\
\bottomrule
\end{tabular}
\end{table}

\subsection{Analysis of Failure Cases}

\begin{figure*} [tb]
    \centering
    \includegraphics[width=\linewidth]{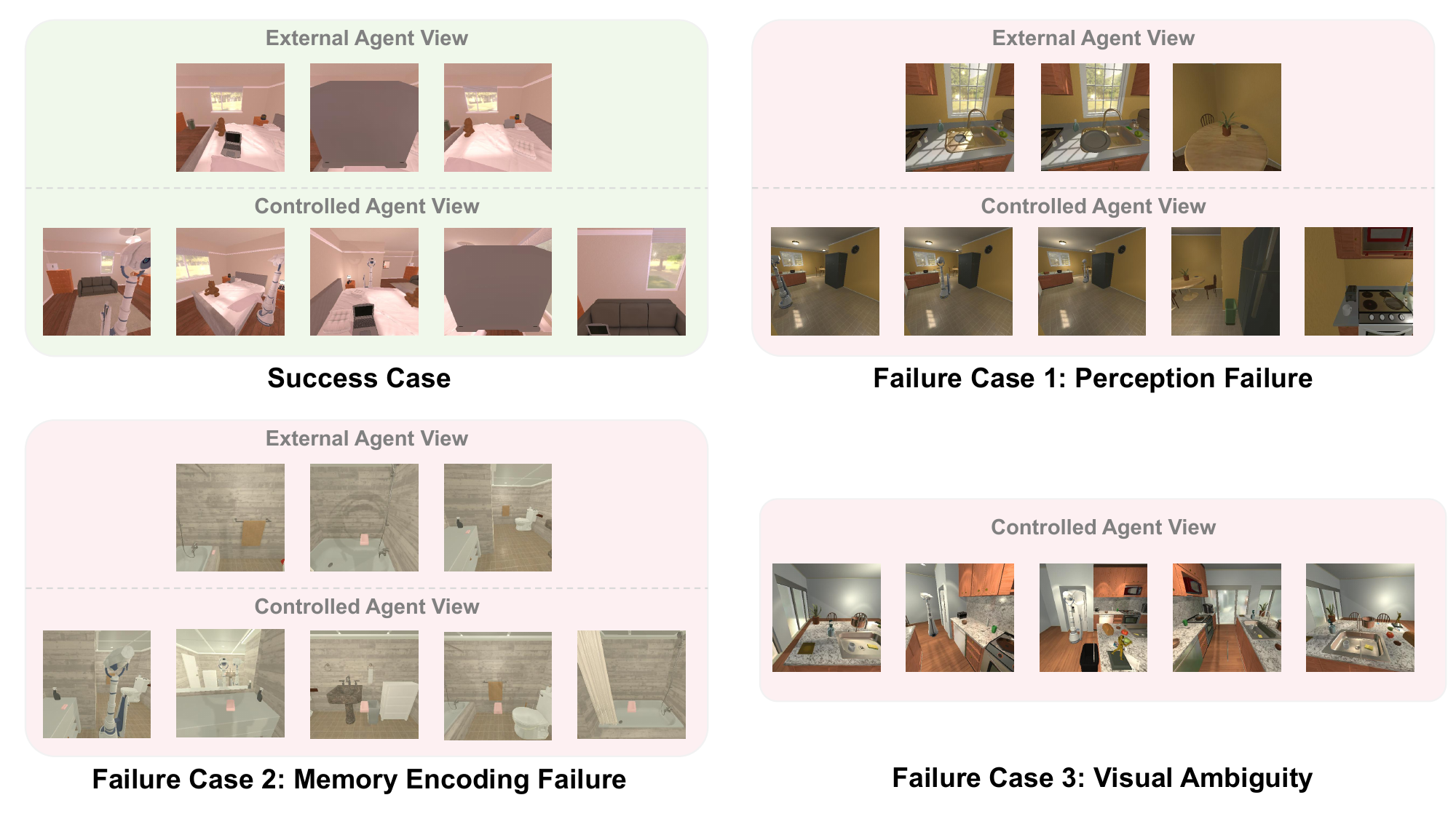}
    \caption{\textbf{Failure cases of task executions on the Dyna-THOR benchmark.} We also illustrate a success case highlighting the agent's ability to adapt to dynamic object displacement through continuous observation. Furthermore, we analyze three representative failure modes: (1) \textbf{Perception Failure} (top-right), where slender object profiles impede recognition; (2) \textbf{Memory Encoding Failure} (bottom-left), causing ineffective exploration due to missed detections; and (3) \textbf{Visual Ambiguity} (bottom-right), arising from semantic confusion between similar objects (e.g., Cup vs. Mug).}
    \label{fig:sup_failures}
\end{figure*}

Fig.~\ref{fig:sup_failures} illustrates representative successful and failed episodes observed during our experiments. In this section, we provide a detailed qualitative analysis of these cases to elucidate the underlying causes of failure.

Successful task execution predicates on the ego-agent's ability to accurately perceive environmental dynamics, specifically object displacements introduced by external agents. As demonstrated in the Success Case in Fig.~\ref{fig:sup_failures}, the external agent moves a Laptop from the bed to a new location. In a standard successful run, the ego-agent initiates retrieval based on historical memory. During navigation, through continuous observation, the ego-agent detects the significant displacement and updates the corresponding memory entry using real-time perception. This mechanism successfully guides the ego-agent to the laptop's new location for subsequent manipulation.

However, failures persist in certain scenarios. We summarize three representative failure modes analyzed from the experiments:

\textbf{Failure Case 1: Perception Failure.} As illustrated in the top-right panel of Fig.~\ref{fig:sup_failures}, the task required placing a \textit{Plate} into the \textit{Sink}, and the external agent had displaced it to the \textit{Dining Table}. While the ego-agent correctly inferred the approximate displacement direction from external agent's trajectory and initiated exploration, the failure occurred at the recognition stage. The target object (\textit{Plate}) exhibits an extremely slender geometric profile. Despite being within the ego-agent's field of view, the visual recognition model failed to identify and segment it due to these subtle geometric features. Consequently, no valid memory entry was generated. This absence forced the system to erroneously retrieve another object with feature correlations to the concept of a ``plate'',ultimately resulting in localization errors.

\textbf{Failure Case 2: Memory Encoding Failure.} In the scenario depicted in the bottom-left panel of Fig.~\ref{fig:sup_failures}, the objective was to dispose of a \textit{SoapBar} into a \textit{GarbageCan}. However, the perception module failed to detect the \textit{GarbageCan} (i.e., failing to generate valid bounding boxes or segmentation masks). As a result, the memory module could not encode a valid entry for the target object. Deprived of a retrieval index for the \textit{GarbageCan}, the agent engaged in prolonged and futile exploration, ultimately failing to locate the target.

\textbf{Failure Case 3: Visual Ambiguity.} This failure arises from visually similar distractors in the environment, which introduce ambiguity during visual matching. In the bottom-right panel of Fig.~\ref{fig:sup_failures}, the task required placing a \textit{Knife} into a \textit{Cup}. The environment contained two candidate objects: a green cylindrical \textit{Cup} on the countertop (the true target) and a white \textit{Mug} in the sink (the distractor). Semantically, a \textit{Mug} is a hyponym of \textit{Cup}. Since the instruction specified ``Cup'' without fine-grained discriminative attributes (e.g., color), the model prioritized the distractor \textit{Mug} due to higher semantic affinity. This misalignment led the agent to execute actions targeting the wrong object, causing task failure due to incorrect localization.

\subsection{More Details in Experiments}

We standardized the agent's ego-centric view resolution to $400 \times 400$ pixels. This setting was calibrated to balance computational efficiency with the inference accuracy of visual perception models (recognition and segmentation). Specifically, this resolution preserves critical geometric features of fine-grained objects, ensuring accurate semantic segmentation while mitigating the computational overhead of higher resolutions, thus guaranteeing real-time system responsiveness.

The core parameters in our framework are set as follows: the semantic threshold $\theta_{sem}$=0.6, the support threshold $\theta_{sup}$=0.5, the support score weights $\omega_1$=0.35, $\omega_2$=0.35, $\omega_3$=0.2, and $\omega_4$=0.1, the Bidirectional ICP thresholds $\sigma_{fwd}$=0.6 and $\sigma_{bwd}$=0.3, the baseline decay rate $\lambda_{base}$=0.1, the retrieval weights $\alpha_1$=1.0 and $\alpha_2$=1.0, and the exploration sector angle $\theta_R$=$\pi$. These values were determined through a combination of references to established practices in prior research and our own empirical tuning.

To accommodate methods requiring environmental priors for initial episodic memory, we implemented a standardized pre-exploration phase. In this phase, the ego-agent autonomously navigates using a Fast Frontier-based Exploration algorithm grounded in a real-time occupancy grid map. By prioritizing paths to the boundaries between explored and unexplored regions, the algorithm maximizes coverage of the spatial structure and object distribution with minimal steps, facilitating the rapid construction of a global semantic memory. Upon completion, the agent returns to its initial spawn pose. This reset mechanism is crucial for eliminating interference caused by the stochastic nature of exploration endpoints, ensuring that all evaluation episodes commence from a consistent, deterministic state. Notably, steps incurred during this pre-exploration phase are explicitly included in the total step cost for statistical analysis.

Complementing the comparative results presented in the main text, our experimental results across four scene categories (kitchen, livingroom, bedroom, bathroom) indicate that the method achieves optimal comprehensive performance in Bedroom scenarios, whereas performance in Kitchen environments is relatively weaker. Our analysis attributes this discrepancy to significant variations in object categories, scales, and shapes across scenes, which differentially affect the robustness of the visual recognition and segmentation models. Specifically, objects in Bedroom scenes (e.g., AlarmClock, Pillow, Book) generally exhibit distinct inter-class visual features and moderate scales, enabling the integrated visual model to generate accurate recognitions and feature representations. In contrast, Kitchen environments present substantial challenges. First, they contain numerous small-scale objects, and certain categories share high visual similarity (e.g., PepperShaker vs. SaltShaker), leading to frequent matching and manipulation errors. Second, Kitchen scenes are characterized by dense object arrangements and occlusion, which compromise the output of the visual model and result in a lower task success rate.

\subsection{Limitations}
We acknowledge several limitations in the current methodology and benchmark design: (1) Dependency on Foundation Models: The accuracy and efficiency of memory construction are fundamentally constrained by the capabilities of the underlying foundation vision models. Perception errors from these models inevitably propagate to the memory module. (2) Lack of Planning Verification: Current task planning relies on the comprehension and generative capabilities of LLMs. However, the existing framework performs verification only during the execution phase, lacking a mechanism for a priori feasibility checking or closed-loop validation of generated action sequences. This absence of pre-execution scrutiny renders the system susceptible to planning hallucinations, where logically invalid or physically impossible actions are proposed without immediate detection. (3) Benchmark Comprehensiveness: The scope of the Dyna-THOR benchmark requires further expansion. Specifically, the diversity of task types needs to be enriched to encompass a broader spectrum of dynamic interaction scenarios.

\section{Algorithm of Retrieval}
\label{sec:supp_algorithm}

We provide the algorithmic implementation of the multi-modal context-aware retrieval mechanism introduced in the main manuscript. To complement the conceptual formulation, Algorithm \ref{alg:retrieval} details the step-by-step execution pipeline. The complete procedure takes the user instruction, the current memory graph $\mathcal{G}_t$, and the historical external agent trajectory $\mathcal{T}$ as inputs, and routes the execution through three core phases:

\begin{algorithm}
\caption{HitMem: Multi-Modal Context-Aware Retrieval}
\label{alg:retrieval}
\textbf{Input}: Instruction $\mathcal{I}$, Memory Graph $\mathcal{G}_t=(\mathcal{V}_t, \mathcal{E}_t)$, Trajectory $\mathcal{T}$, Threshold $\theta_{target}$ \\
\textbf{Output}: Target node $v_{target}$ or Explore
\begin{algorithmic}[1]
\State $F_{query}, Cat_t \gets \text{Parse}(\mathcal{I})$
\State $\mathcal{C}_r, \mathcal{C}_d \gets \text{HierarchicalFilter}(\mathcal{G}_t)$

\State \Comment{\textit{1. Hierarchical Semantic Querying}}
\State $\mathcal{S} \gets \{ \alpha_1 \text{sim}(F_i, F_{query}) + \alpha_2 \mathcal{A}(v_i, t) \mid \forall v_i \in \mathcal{C}_d \}$
\State $\mathcal{V}_{top} \gets \text{TopK}(\{v_i \in \mathcal{C}_d \mid \mathcal{S}[i] > \theta_{target}\}, 3)$

\State \Comment{\textit{2. Active Verification and Pruning}}
\For{$v_i \in \mathcal{V}_{top}$ \textbf{in descending order}}
    \State $obs \gets \text{NavigateAndObserve}(pos(v_i))$
    \If{$obs \equiv Cat_t$}
        \State \Return $v_i$
    \Else
        \State $\mathcal{V}_t \gets \mathcal{V}_t \setminus \{v_i\}$ \Comment{Conflict: Prune obsolete node}
        \State \textbf{break} \Comment{Trigger relocation}
    \EndIf
\EndFor

\State \Comment{\textit{3. Two-Stage Relocation}}
\State \textit{// Stage 1: Search Space Filtering}
\State $c \gets \arg\min_{i} \| \mathcal{T}[i] - pos_{last} \|_2$ \Comment{Index of $\mathcal{P}_{closest}$}
\State $\tau_{local} \gets \mathcal{T}[c : c+k]$
\State $\vec{D} \gets \sum_{m=1}^{k} \Delta\mathcal{A}_{c+m} \cdot \Delta pos_{c+m}$ \Comment{Ref. Eq. 1}
\State $\mathcal{C}_{cand} \gets \left\{ v \in \mathcal{C}_r \;\middle|\; \text{cos\_sim}(\vec{D}, pos_v - pos_{last}) \ge \cos(\frac{\theta_R}{2}) \right\}$ \Comment{Ref. Eq. 2}

\Statex
\State \textit{// Stage 2: Class Affinity Scanning}
\State $\mathcal{P}_{explore} \gets \text{Softmax}(\{ A_{affinity}(Cat_t, v_j) \mid \forall v_j \in \mathcal{C}_{cand} \})$ \Comment{Ref. Eq. 3}
\For{$v_j \in \text{SortDescending}(\mathcal{C}_{cand}, \mathcal{P}_{explore})$}
    \State $obs \gets \text{NavigateAndObserve}(pos(v_j))$
    \If{$obs \equiv Cat_t$}
        \State \Return $v_{new}$
    \EndIf
\EndFor
\State \Return Explore
\end{algorithmic}
\end{algorithm}

\begin{itemize}
    \item \textbf{Hierarchical Semantic Querying}: Efficiently narrows down candidate nodes within $\mathcal{G}_t$ by fusing semantic relevance and temporal recency.
    \item \textbf{Active Verification \& Pruning}: Physically navigates to the anticipated location to verify the target, proactively purging obsolete graph nodes upon detecting a memory-observation conflict.
    \item \textbf{Two-Stage Relocation}: Systematically re-localizes displaced objects by first bounding the physical search space via the external agent's motion vector, followed by a probabilistic exploration guided by class affinity.
\end{itemize}

\section{Prompts of LLM Planning}

\begin{figure*} [htbp]
    \centering
    \includegraphics[width=0.94\linewidth]{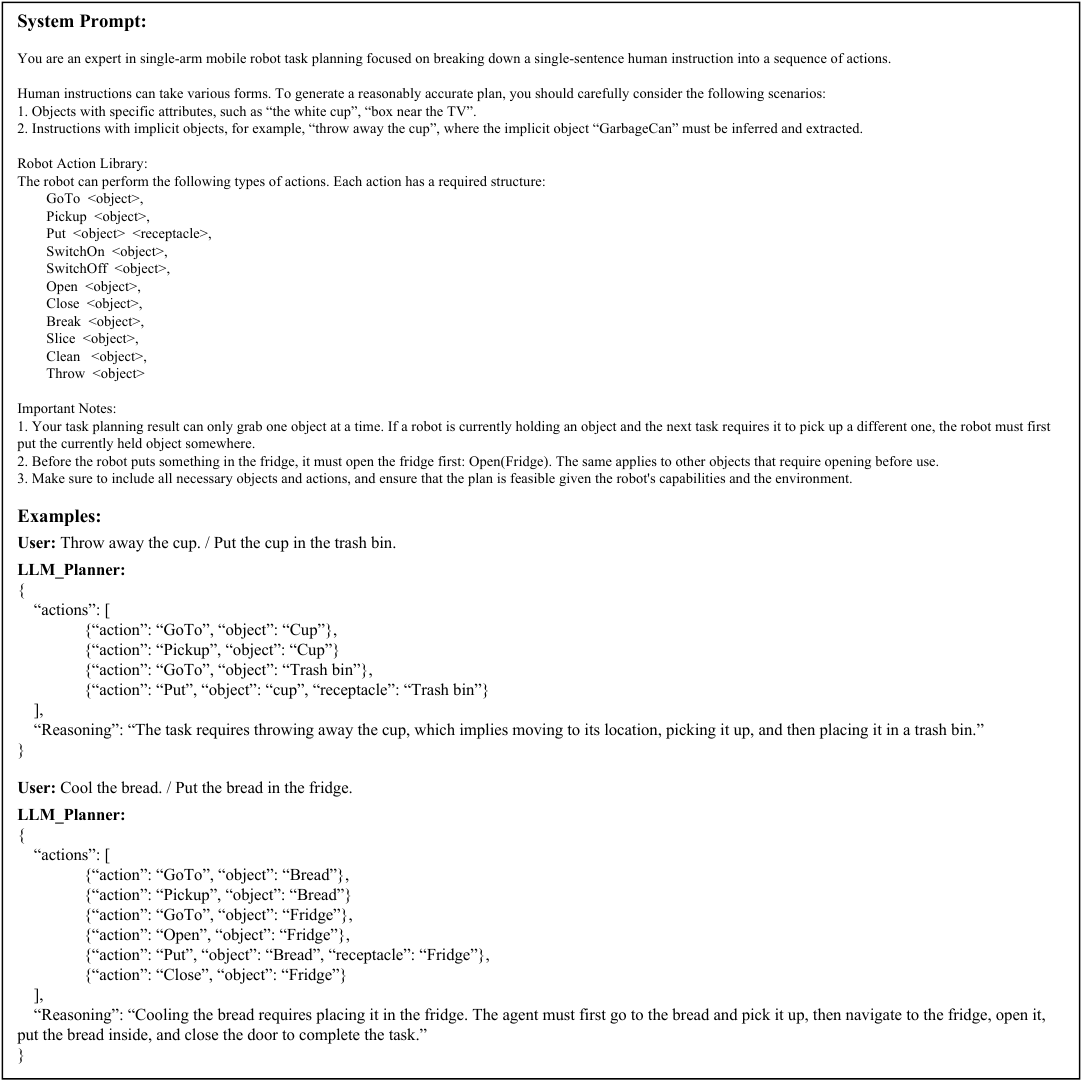}
    \caption{Prompt for task planning.}
    \label{fig:sup_prompt}
\end{figure*}

Fig.~\ref{fig:sup_prompt} illustrates the prompt design employed for LLM-based task planning. To facilitate this process, we utilize a few-shot prompting strategy, embedding representative planning examples to serve as references.

\end{document}